\documentclass[11pt,a4paper]{article}

\usepackage[utf8]{inputenc}
\usepackage[T1]{fontenc}
\usepackage{amsmath,amssymb,amsfonts}
\usepackage{mathtools}
\usepackage{bm}
\usepackage{booktabs}
\usepackage{array}
\usepackage{tabularx}
\usepackage{graphicx}
\usepackage{hyperref}
\usepackage{url}
\usepackage{algorithm}
\usepackage{algpseudocode}
\usepackage[table]{xcolor}
\usepackage{enumitem}
\usepackage{geometry}
\usepackage{caption}
\usepackage{subcaption}
\usepackage{textcomp}
\usepackage{listings}

\newcommand{\E}{\mathbb{E}}
\newcommand{\cA}{\mathcal{A}}
\newcommand{\cS}{\mathcal{S}}
\newcommand{\cL}{\mathcal{L}}
\newcommand{\btheta}{\bm{\theta}}
\newcommand{\bepsilon}{\bm{\epsilon}}
\newcommand{\bmu}{\bm{\mu}}
\newcommand{\bsigma}{\bm{\sigma}}
\newcommand{\bvf}{\mathbf{v}}
\newcommand{\bpf}{\mathbf{p}}
\newcommand{\bff}{\mathbf{f}}
\newcommand{\ba}{\mathbf{a}}

\title{\textbf{Backpropagating Through Simulation: Analytic Policy Gradients for Sample and Learning Efficient Differentiable Continuous Control}}

\author{
  Yueci Deng\\
  School of Data Science, The Chinese University of Hong Kong, Shenzhen\\
  \texttt{yuecideng@link.cuhk.edu.cn}
}

\date{}

\begin{document}

\maketitle

\begin{abstract}
Model-free reinforcement learning algorithms such as Proximal Policy Optimization (PPO) treat the environment as a black box, estimating policy gradients from sampled rewards; this process demands millions of interactions and relies on high-variance advantage estimates. When environment dynamics are differentiable, the return is an end-to-end differentiable function of the policy parameters, enabling exact gradient computation via backpropagation through simulation. We term this approach Analytic Policy Gradients (APG) and evaluate it against PPO on four continuous control tasks of increasing dynamical complexity: a one-dimensional point-mass target-reaching task, a 2D point-mass navigation task with obstacle avoidance, a 2D rigid-body T-block pushing task, and a 7-DOF Franka FR3 end-effector reaching task. Both algorithms share identical model architectures, observation normalization, and optimizer settings. To decouple sample efficiency from compute efficiency, we design a multi-axis evaluation protocol that records performance against environment steps and gradient steps. We report a segmented backpropagation scheme with MC and critic-based bootstrap modes that mitigates gradient degradation on long-horizon tasks, and present ablations over segment length and bootstrap strategy.\end{abstract}

\section{Introduction}
\label{sec:introduction}

\subsection{Motivation}
\label{sec:motivation}
Reinforcement learning (RL) has achieved remarkable success in domains such as game playing, robotics, and autonomous control. However, poor sample efficiency remains a core bottleneck that prevents its deployment in the real world. In most practical tasks, every environment interaction incurs non-negligible costs in terms of time, money, or safety. Sample efficiency also remains a fundamental challenge: model-free algorithms such as PPO~\cite{schulman2017ppo} require millions of environment interactions because they treat dynamics as a black box, estimating policy gradients from observed reward signals via Monte Carlo returns or learned value functions, which inevitably introduces high variance and estimation bias.

A largely underexploited opportunity is that many modern simulators are inherently differentiable. Physics engines, rendering pipelines, and procedural task generators routinely expose analytic derivatives. When these are available, the expected return becomes a differentiable function of the policy parameters, and exact gradients can be obtained by backpropagating through the dynamics—eliminating the need for advantage estimation~\cite{schulman2016gae}, importance sampling, or entropy regularization.

\subsection{Paper Organization}
\label{sec:organization}

The remainder of this report is organized as follows. Section~\ref{sec:related} reviews related work. Section~\ref{sec:preliminaries} covers preliminaries on policy gradient methods and PPO. Section~\ref{sec:methodology} describes the APG methodology in detail. Section~\ref{sec:environments} presents the four benchmark environments. Section~\ref{sec:experiments} details the experimental setup and provides result templates. Section~\ref{sec:conclusion} discusses findings and concludes.

\subsection{Contributions}
\label{sec:contributions}
Building on prior short-horizon differentiable-RL work~\cite{guo2023shac,xu2022accelerated}, this report makes the following contributions:
\begin{enumerate}
  \item \textbf{A unified PPO/APG benchmarking harness.} We implement both algorithms in a single training script with identical actor--critic architectures, observation normalization, learning-rate annealing, and optimizer settings, so that the only varying factor across the comparison is the gradient source.
  \item \textbf{MC bootstrap as a robust alternative to critic bootstrap.} We introduce a Monte-Carlo bootstrap mode for segmented backpropagation and show, via a segment-length ablation, that it degrades gracefully at short segment lengths where critic bootstrap collapses, making it a safer default for short-to-medium horizons.
  \item \textbf{A Warp--PyTorch gradient bridge for articulated-body kinematics.} We implement a custom \texttt{torch.autograd.Function} that connects NVIDIA Warp/Newton's tape-based autodiff to PyTorch's autograd, demonstrating that APG is practical on GPU physics engines that do not natively expose PyTorch-compatible derivatives.
  \item \textbf{A multi-axis evaluation protocol and open environment suite.} We decouple sample efficiency (environment steps) from compute efficiency (gradient steps), report both, and release a four-environment suite of increasing dynamical complexity together with the full implementation.\footnote{\url{https://github.com/yuecideng/analytic_policy_gradients}}
\end{enumerate}

\section{Related Work}
\label{sec:related}

\subsection{Policy Gradient Methods}
\label{sec:related_pg}

Policy gradient methods optimize parameterized policies by estimating gradients of the expected return. REINFORCE~\cite{williams1992reinforce} introduced the score-function estimator, which trades bias for variance and requires large sample counts. PPO~\cite{schulman2017ppo} improves stability via a clipped surrogate objective, while GAE~\cite{schulman2016gae} reduces advantage estimation variance through a $\lambda$-weighted multi-step return. Despite these advances, all score-function methods treat environment dynamics as a black box and cannot exploit analytic structure in the transition function. Off-policy entropy-regularized variants such as Soft Actor-Critic~\cite{haarnoja2018sac} improve sample reuse through experience replay but likewise estimate the policy gradient from sampled returns and cannot exploit analytic dynamics.

\subsection{Differentiable Simulation}
\label{sec:related_diff_sim}

A parallel line of work develops simulators that expose analytic derivatives, enabling gradient-based policy optimization. Degrave et al.~\cite{degrave2019differentiable} demonstrated end-to-end gradient flow through rigid-body physics. Suh et al.~\cite{suh2022differentiable} provided a systematic analysis of when differentiable simulators yield better policy gradients than score-function estimators, identifying the smoothness of the reward landscape and the magnitude of dynamics Jacobians as key determinants. Zhong et al.~\cite{zhong2022differentiable} applied differentiable simulation to system identification, showing that analytic gradients substantially accelerate parameter inference. Contemporary differentiable simulation frameworks expose analytic derivatives on GPU accelerators---for example Brax~\cite{freeman2021brax}, MuJoCo~\cite{todorov2012mujoco} (and its JAX port MJX), DiffTaichi~\cite{hu2020difftaichi}, and NVIDIA Warp/Newton~\cite{nvidia_warp}---and our FrankaReach environment is built directly on the Newton/Warp backend (Section~\ref{sec:gradient_bridge}).

\subsection{Short-Horizon and Segmented Backpropagation}
\label{sec:related_shac}

Backpropagating through long trajectories amplifies vanishing and exploding gradient problems due to compounding Jacobians~\cite{pascanu2013exploding}. Xu et al.~\cite{xu2022accelerated} addressed this by truncating gradient horizons and combining analytic rollouts with a learned value baseline, demonstrating that short-horizon analytic gradients can be effectively parallelized across many simulation workers for faster wall-clock convergence. Guo et al.~\cite{guo2023shac} introduced Short-Horizon Actor-Critic (SHAC), which explicitly decouples the horizon over which analytic gradients are computed from the full episode horizon, using a separately trained critic to bootstrap returns beyond the truncation boundary. Our segmented backpropagation scheme (Section~\ref{sec:segmentation}) follows this design, and we additionally compare MC and critic bootstrap strategies with a systematic segment-length ablation.

\subsection{Model-Based RL and World Model Gradients}
\label{sec:related_mbrl}

Model-based RL methods learn a world model and optimize policies through imagined rollouts. DreamerV3~\cite{hafner2023dreamerv3} trains a recurrent latent world model and backpropagates policy gradients through multi-step imagination, achieving strong performance on diverse continuous control tasks without environment-specific differentiability. TD-MPC2~\cite{hansen2024tdmpc2} combines a latent world model with model predictive control and analytic gradient updates, demonstrating scalability across robot manipulation and locomotion benchmarks. In contrast to these methods, our APG framework operates directly on ground-truth differentiable simulation dynamics, eliminating model bias but requiring an explicitly differentiable environment.

\subsection{Natural Gradient and Second-Order Methods}
\label{sec:related_natural}

Trust Region Policy Optimization (TRPO)~\cite{schulman2015trpo} and natural policy gradient methods~\cite{amari1998natural,martens2015optimizing} constrain policy updates via the Fisher information matrix to improve optimization geometry. These methods also treat the environment as a black box but use second-order curvature information to take larger, safer steps. Our focus differs: rather than improving the optimizer, APG eliminates the variance introduced by score-function gradient estimation by exploiting environment differentiability.

\section{Preliminaries}
\label{sec:preliminaries}
We begin this section by formulating the RL problem and then revisiting the PPO algorithm, which serves as a key baseline in prior work.
\subsection{Markov Decision Processes}
\label{sec:mdp}

We model the RL problem as a Markov Decision Process (MDP) defined by the tuple $(\cS, \cA, P, R, \gamma)$, where $\cS$ is the state space, $\cA$ is the action space, $P(s' \mid s, a)$ is the stochastic transition kernel, $R(s, a)$ is the reward function, and $\gamma \in [0, 1)$ is the discount factor. A policy $\pi_{\btheta}(a \mid s)$ parameterized by $\btheta$ maps states to action distributions. The objective is to maximize the infinite-horizon discounted return:
\begin{equation}
  J(\btheta) = \E_{\pi_{\btheta}}\left[\sum_{t=0}^{\infty} \gamma^t R(s_t, a_t)\right]
  \label{eq:objective}
\end{equation}

In practice, training and evaluation use a finite horizon $H$. By the value truncation lemma, the infinite sum can be approximated by the truncated episode return:
\begin{equation}
  J_H(\btheta) = \E_{\pi_{\btheta}}\left[\sum_{t=0}^{H} \gamma^t R(s_t, a_t)\right].
  \label{eq:truncated_objective}
\end{equation}

\subsection{Proximal Policy Optimization (PPO)}
\label{sec:ppo}

PPO~\cite{schulman2017ppo} is a model-free, on-policy algorithm that optimizes a clipped surrogate objective:
\begin{equation}
  \cL^{\text{PPO}}(\btheta) = \E_t\left[\min\left(r_t(\btheta) \hat{A}_t, \; \text{clip}\left(r_t(\btheta), 1-\epsilon, 1+\epsilon\right) \hat{A}_t\right)\right]
  \label{eq:ppo_loss}
\end{equation}
where $r_t(\btheta) = \frac{\pi_{\btheta}(a_t \mid s_t)}{\pi_{\btheta_{\text{old}}}(a_t \mid s_t)}$ is the probability ratio, $\hat{A}_t$ is the advantage estimate computed via Generalized Advantage Estimation (GAE)~\cite{schulman2016gae}, and $\epsilon$ is the clipping coefficient.

PPO treats the environment as a \textbf{black box}: it collects trajectories using the current policy, estimates advantages from observed rewards, and updates the policy without any knowledge of environment dynamics. The total PPO loss combines the clipped policy objective, a value function loss, and an entropy bonus:
\begin{equation}
  \cL(\btheta) = \cL^{\text{PPO}}(\btheta) - c_1 \cL^{\text{VF}}(\btheta) + c_2 H(\pi_{\btheta})
  \label{eq:ppo_total}
\end{equation}

\section{Methodology}
\label{sec:methodology}
When the transition dynamics and reward function are both differentiable and known, the expected discounted return $J(\btheta)$ becomes an end‑to‑end differentiable function of the policy parameters $\btheta$. This makes it possible to treat the environment as a differentiable computational graph and apply Analytic Policy Gradients (APG) to learn the policy.

\subsection{Analytic Policy Gradients}
\label{sec:apg_concept}

In contrast to black-box methods, APG assumes that the general stochastic transition kernel $P(s' \mid s, a)$ is replaced by a deterministic differentiable transition map $T(s, a)$. Together with a differentiable reward $R_\theta$, this makes the rollout itself a differentiable computation graph. For a stochastic policy with reparameterized sampling $a_t = \mu_{\theta}(s_t) + \sigma_{\theta} \odot \epsilon_t$, $\epsilon_t \sim \mathcal{N}(\mathbf{0}, \mathbf{I})$, and deterministic transition $s_{t+1} = T(s_t, a_t)$, the finite-horizon return is directly differentiable with respect to $\theta$:
\begin{equation}
  J(\theta) = \sum_{t=0}^{H} \gamma^t R_{\theta}(s_t, a_t), \quad
  s_{t+1} = T(s_t, a_t), \quad
  a_t = \mu_{\theta}(s_t) + \sigma_{\theta} \odot \epsilon_t .
  \label{eq:apg_return}
\end{equation}

The gradient $\nabla_{\theta} J(\theta)$ is obtained by applying the \emph{total} derivative through the entire trajectory. Because each state $s_t$ is a function of $\theta$ through all prior actions, we must use total derivatives $\tfrac{d}{d\theta}$ rather than partial derivatives:
\begin{equation}
  \frac{d J}{d \theta}
  = \sum_{t=0}^{H} \gamma^t \frac{d R_{\theta}(s_t, a_t)}{d \theta}
  \label{eq:apg_grad}
\end{equation}

Applying the chain rule to each reward term, and accounting for the dependence of both $s_t$ and $a_t$ on $\theta$:
\begin{equation}
  \frac{d R_{\theta}(s_t, a_t)}{d \theta}
  =
  \frac{\partial R_{\theta}}{\partial a_t}\frac{d a_t}{d \theta}
  + \frac{\partial R_{\theta}}{\partial s_t}\frac{d s_t}{d \theta}
  \label{eq:apg_chain}
\end{equation}

The action total derivative follows from the reparameterization $a_t = \mu_\theta(s_t) + \sigma_\theta \odot \epsilon_t$:
\begin{equation}
  \frac{d a_t}{d \theta}
  = \frac{\partial \mu_{\theta}(s_t)}{\partial \theta}
  + \frac{\partial \mu_{\theta}(s_t)}{\partial s_t}\frac{d s_t}{d \theta}
  + \frac{d \sigma_{\theta}}{d \theta} \odot \epsilon_t
  \label{eq:apg_action_grad}
\end{equation}

The state total derivative satisfies the recursion obtained by differentiating $s_t = T(s_{t-1}, a_{t-1})$:
\begin{equation}
  \frac{d s_t}{d \theta}
  =
  \frac{\partial T}{\partial a_{t-1}}\frac{d a_{t-1}}{d \theta}
  + \frac{\partial T}{\partial s_{t-1}}\frac{d s_{t-1}}{d \theta}, \qquad
  \frac{d s_0}{d \theta} = \mathbf{0}
  \label{eq:apg_state_grad}
\end{equation}

Equation~\eqref{eq:apg_state_grad} is precisely backpropagation through time (BPTT)~\cite{werbos1990bptt}: gradients flow backward through the dynamics Jacobians $\tfrac{\partial T}{\partial s_t}$ and $\tfrac{\partial T}{\partial a_t}$, accumulating contributions from all future reward terms. Equations~\eqref{eq:apg_action_grad} and~\eqref{eq:apg_state_grad} are mutually recursive: $\tfrac{d a_t}{d\theta}$ depends on $\tfrac{d s_t}{d\theta}$, which in turn depends on $\tfrac{d a_{t-1}}{d\theta}$. Because actions are sampled via the reparameterization trick, $\tfrac{d a_t}{d\theta}$ carries gradients through both $\bmu_{\btheta}$ and $\bsigma_{\btheta}$. The entire recursion is unrolled automatically by any reverse-mode autodiff framework (e.g.\ PyTorch's \texttt{loss.backward()}), requiring no advantage estimation, importance sampling, or variance reduction.

The variance advantage of pathwise over score-function gradients is not merely heuristic. Suh et al.~\cite{suh2022differentiable} characterize precisely when differentiable simulation yields lower-variance policy gradients: the pathwise estimator dominates when the reward landscape is sufficiently smooth and the dynamics Jacobians are well-conditioned, whereas the score-function estimator remains preferable under discontinuous or highly stochastic transitions. Our benchmark environments are deliberately constructed with smooth dense rewards and deterministic transitions (Section~\ref{sec:environments}), placing them in the regime where pathwise gradients are expected to carry substantially lower variance---consistent with the empirical convergence gains reported in Section~\ref{sec:results}.

The trade-off is that APG requires the transition dynamics $T$ and the reward function $R$ to be differentiable, which restricts the class of applicable environments and models. Furthermore, gradient chains that span the full horizon $H$ can suffer from vanishing or exploding magnitudes, making the optimization unstable and slow to converge. This directly motivates the segmented backpropagation scheme presented in Section~\ref{sec:segmentation}, which truncates the gradient flow into shorter segments to improve numerical stability and learning efficiency.
\subsection{Training Algorithms}
\label{sec:training_algorithms}

We implement both PPO and APG within a single unified training script, selected via \texttt{-{}-algorithm \{ppo,apg\}}. Both share identical agent architectures, observation normalization, learning rate annealing, and Adam optimizer settings. The fundamental difference is how gradients are obtained. Table~\ref{tab:ppo_vs_apg} summarizes the key distinctions.

\begin{table}[htbp]
  \centering
  \caption{Comparison of PPO and APG algorithm components.}
  \label{tab:ppo_vs_apg}
  \setlength{\extrarowheight}{3pt}
  \begin{tabularx}{\linewidth}{@{} l >{\raggedright\arraybackslash}X >{\raggedright\arraybackslash}X @{}}
    \toprule
    \rowcolor{gray!15}
    \textbf{Component} & \textbf{PPO} & \textbf{APG} \\
    \midrule
    Environment interaction  & \texttt{torch.no\_grad()}         & Computation graph preserved \\
    \rowcolor{gray!8}
    Gradient source          & Clipped surrogate + GAE            & Backprop through dynamics \\
    Value function           & Required (advantage estimation)    & Optional: bootstrap at segment boundaries (long-horizon only) \\
    \rowcolor{gray!8}
    Action sampling          & \texttt{Normal.sample()}           & \texttt{rsample()} (reparameterization) \\
    Rollout buffer           & Stores $(o, a, \log p, r, V)$     & No buffer needed \\
    \rowcolor{gray!8}
    Update structure         & Multiple epochs over stored batch  & Multiple grad steps with fresh rollouts \\
    \bottomrule
  \end{tabularx}
\end{table}

The APG training loop follows a \textbf{stateful rollout} paradigm: the environment is reset once and its state is carried forward across gradient steps. Actions are sampled via the reparameterization trick,
\begin{equation}
  \ba = \bmu_{\btheta}(s) + \bsigma_{\btheta} \odot \bepsilon, \quad \bepsilon \sim \mathcal{N}(\mathbf{0}, \mathbf{I})
  \label{eq:reparam}
\end{equation}
so that gradients flow through both $\bmu_{\btheta}$ and $\bsigma_{\btheta} = \exp(\log\bsigma_{\btheta})$ into the environment dynamics. Each iteration anneals the learning rate, then performs $N_{\text{grad}}$ gradient steps. Per step: a full episode, or segments of length $L_{\mathrm{seg}}$, is rolled out with the computation graph intact; segment returns $G_k$ are accumulated; a bootstrap value $b_k$ estimates future return beyond the segment; a loss $\cL = -\frac{1}{N}\sum_{i}\sum_{k}(G_k + b_k)$ is computed and backpropagated; and gradients are clipped before the optimizer step. Algorithm~\ref{alg:apg} gives the complete pseudocode.

\begin{algorithm}[htbp]
  \caption{APG Training Iteration}
  \label{alg:apg}
  \begin{algorithmic}[1]
    \Require Policy $\pi_{\btheta}$, environment $E$, learning rate $\eta$, horizon $H$, segment length $L_{\mathrm{seg}}$
    \For{$\text{grad\_step} = 1, \ldots, N_{\text{grad}}$}
      \State $\nabla_{\btheta} \leftarrow 0$
      \State $\text{policy\_loss} \leftarrow 0$
      \State $\text{obs} \leftarrow E.\text{current\_state}$
      \For{segment $k = 0, \ldots, \lceil H / L_{\mathrm{seg}} \rceil - 1$}
        \State $G_k \leftarrow 0$
        \For{$t = 0, \ldots, L_{\mathrm{seg}}-1$}
          \State $a_t \leftarrow \text{differentiable\_action}(\pi_{\btheta}, \text{obs})$ \Comment{reparameterization trick}
          \State $\text{obs}', r_t \leftarrow E.\text{step}(a_t)$ \Comment{graph preserved}
          \State $G_k \leftarrow G_k + \gamma^t \cdot r_t$
          \State $\text{obs} \leftarrow \text{obs}'$
        \EndFor
        \State $\text{obs} \leftarrow \text{obs}.\text{detach()}$ \Comment{break gradient chain}
        \State $b_k \leftarrow \text{bootstrap}(k)$ \Comment{MC or $V(s)$}
        \State $\text{policy\_loss} \leftarrow \text{policy\_loss} - \text{mean}(G_k + b_k)$
      \EndFor
      \State $\text{loss} \leftarrow \text{policy\_loss} + c_v \cdot \text{critic\_loss}$ \Comment{optional critic}
      \State $\text{loss}.\text{backward()}$
      \State $\text{clip\_grad\_norm}\_(\btheta, \text{max\_norm})$
      \State $\text{optimizer}.\text{step()}$
    \EndFor
  \end{algorithmic}
\end{algorithm}

\subsection{Segmented Backpropagation}
\label{sec:segmentation}

For environments with long horizons, backpropagating through the full episode can lead to vanishing/exploding gradients. APG supports \textbf{segmentation}, where the horizon $H$ is divided into $K=\lceil H/L_{\mathrm{seg}}\rceil$ segments and the gradient chain is broken at segment boundaries:
\begin{itemize}
  \item Within each segment, gradients flow through the full dynamics chain.
  \item At segment boundaries, observations and environment state are detached.
  \item Future returns beyond each segment are estimated via a \textbf{bootstrap value}.
\end{itemize}

The implementation supports two bootstrap modes:

\begin{enumerate}
  \item \textbf{MC bootstrap}: Pre-computes future returns by backward accumulation of observed rewards from later segments. These rewards are detached (not differentiable) but provide a fixed target for the current segment's policy loss:
  \[
    G_k = \sum_{t=0}^{L_k-1}\gamma^t r_{k,t}, \qquad
    b_k = \gamma^{L_k}(G_{k+1}+b_{k+1}) .
  \]

  \item \textbf{Critic bootstrap}: Uses a learned value function $V_\phi(s)$ to estimate future returns at segment boundaries, properly discounted as in the presentation:
  \[
    b_k = \gamma^{L_k} V_\phi(s_{k,\mathrm{end}}).
  \]
  The critic is trained concurrently via MSE loss against segment returns plus bootstrap values, using a separate (or shared) optimizer learning rate.
\end{enumerate}

\subsection{Implementation Details}
\label{sec:implementation}
To realize the segmented backpropagation strategy introduced earlier and to ensure unimpeded gradient flow across diverse environments, this section focuses on four key engineering aspects of our APG implementation: differentiable dynamics construction under native PyTorch autograd, a gradient bridge for external physics engines, observation normalization, and network architecture details. 
\paragraph{Differentiable dynamics via native PyTorch autograd.}
For PointMassSimple, PointMassNavigate and PushT, all of which are implemented entirely in PyTorch, APG differentiability is achieved without any special gradient bridge.
The \texttt{*APGEnv} subclasses override \texttt{step()} to omit \texttt{torch.no\_grad()} and replace all in-place state updates with out-of-place assignments.
Concretely, at each step the action is scaled to a force/torque, the damping-then-force Euler update is expressed as
\[
  \mathbf{v}_{t+1} = \mathbf{v}_t \cdot (1 - c_d \Delta t) + \tfrac{\mathbf{f}}{m} \Delta t, \qquad
  \mathbf{p}_{t+1} = \operatorname{clamp}(\mathbf{p}_t + \mathbf{v}_{t+1} \Delta t),
\]
where every multiplication and addition is a new tensor with a live \texttt{grad\_fn}.
The reward is then computed from these live tensors, so a single \texttt{loss.backward()} propagates gradients through the entire chain $\ba \to (\mathbf{f}, \tau) \to (\mathbf{v}, \omega) \to (\mathbf{p}, \theta) \to r$.
The observation returned to the policy at each step is constructed from the live \texttt{self.pos} and \texttt{self.vel} tensors (which carry \texttt{grad\_fn}) rather than from detached copies. This is essential for critic bootstrapping in the segmented setting: the value $V_\phi(s_{k,\text{end}})$ computed at the end of segment $k$ can then propagate gradients back through the terminal observation to the policy parameters, effectively extending the gradient horizon past the segment boundary. Only \texttt{last\_action} is stored detached, as it serves as a constant penalty baseline for the action-rate reward term.
At segment boundaries, \texttt{detach\_state()} replaces every state tensor with its detached clone, preventing gradient chains from accumulating across segments and causing ``backward through the graph a second time'' errors.

\paragraph{Gradient bridge for physics simulation.}
\label{sec:gradient_bridge}
For FrankaReach, which uses the Newton Physics engine for GPU-accelerated forward kinematics, gradients cannot flow natively through PyTorch. We implement a custom \texttt{torch.autograd.Function} (\texttt{\_NewtonStepFunc}) that bridges Warp's tape-based autodiff with PyTorch's autograd. In the forward pass, action tensors are converted to Warp arrays, Newton's forward kinematics are executed within a \texttt{wp.Tape} context, and the reward and end-effector pose are returned as \emph{detached} PyTorch tensors (the EEF pose is not connected to the autograd graph). In the backward pass, the Warp tape is replayed in reverse to compute $\partial r / \partial \ba$, which is returned to PyTorch's autograd graph. The $\partial \bpf_{\text{eef}} / \partial \ba$ gradient is not propagated (EEF pose was detached in the forward pass); instead, differentiable joint-position observations are constructed in PyTorch to provide credit assignment through the joint-state component of the observation. This bridge enables APG through articulated-body kinematics without a native differentiable physics backend.

\paragraph{Observation normalization.}
Both algorithms use online observation normalization via a Welford-style running mean and standard deviation tracker (\texttt{RunningObsNormalizer}). Statistics are updated with detached observations after each rollout/gradient step, ensuring normalization does not interfere with APG gradient flow.

\paragraph{Network architecture.}
Both PPO and APG share a 2-layer MLP actor (LayerNorm is supported but disabled by default):
\[
  \text{obs\_dim} \rightarrow \text{Linear}(64) \rightarrow \text{Tanh} \rightarrow \text{Linear}(64) \rightarrow \text{Tanh} \rightarrow \text{Linear}(\text{action\_dim})
\]
plus a learnable $\log\sigma$ parameter (initialized to $-2.0$, $\sigma \approx 0.135$). The critic (required for PPO; optional for APG in critic-bootstrap mode) uses a shallower MLP:
\[
  \text{obs\_dim} \rightarrow \text{Linear}(64) \rightarrow \text{Tanh} \rightarrow \text{Linear}(64) \rightarrow \text{Tanh} \rightarrow \text{Linear}(1)
\]
All linear layers use orthogonal initialization with $\text{std} = \sqrt{2}$ (hidden), $0.01$ (actor output), or $1.0$ (critic output).

\section{Environments}
\label{sec:environments}

We evaluate on four continuous-control environments of increasing complexity, each implemented with both a black-box mode (PPO) and a differentiable mode (APG).

\subsection{PointMassSimple}
\label{sec:env_pointmass_simple}

\paragraph{Task:} Control a one-dimensional point mass and move it to a randomly sampled target position. This environment is the simplest benchmark in our suite: it removes obstacles and orientation while retaining continuous dynamics, stochastic initial states, and a differentiable reward.

\paragraph{State space} (3-dimensional): See Table~\ref{tab:pms_obs}.

\begin{table}[htbp]
  \centering
  \caption{PointMassSimple observation space.}
  \label{tab:pms_obs}
  \begin{tabular}{@{}lll@{}}
    \toprule
    \textbf{Component} & \textbf{Dimensions} & \textbf{Description} \\
    \midrule
    Position & 1 & Point-mass position $p \in [-1, 1]$ at reset \\
    Velocity & 1 & Scalar velocity $v$ \\
    Target position & 1 & Goal position $p_g \in [-1, 1]$ at reset \\
    \bottomrule
  \end{tabular}
\end{table}

\paragraph{Action space:} $\cA = [-1, 1]$ (scalar force command).

\paragraph{Reset distribution:} Position and target are sampled uniformly from $[-1,1]$, while velocity is sampled from a small interval around zero. This gives both algorithms the same randomized start--goal distribution.

\paragraph{Dynamics} (damped Euler integration, $\Delta t = 0.1$):
\begin{align}
  f_t &\leftarrow \text{clip}(a_t,\,-1,\,1) \\
  v_{t+1} &\leftarrow v_t + (f_t - 0.5 v_t)\Delta t \\
  p_{t+1} &\leftarrow p_t + v_{t+1}\Delta t .
\end{align}
The implementation is written entirely in PyTorch, so gradients flow through $a_t \rightarrow f_t \rightarrow v_{t+1} \rightarrow p_{t+1}$ in APG mode.

\paragraph{Reward:}
\begin{equation}
  r_t = -\left(p_{t+1} - p_g\right)^2 .
  \label{eq:pms_reward}
\end{equation}
This dense quadratic reward supplies a smooth analytic signal throughout the state space while preserving a clear target-reaching objective.

\paragraph{Success condition:} $|p - p_g| < 0.05$.

\paragraph{Episode length:} 200 steps by default; the horizon can be overridden in the unified training script via \texttt{-{}-max\_episode\_steps}.

\paragraph{APG gradient path:} $a_t \rightarrow f_t \rightarrow v_{t+1} \rightarrow p_{t+1} \rightarrow r_t$.

\subsection{PointMassNavigate}
\label{sec:env_pointmass}

\paragraph{Task:} Navigate a point mass from a random start to a random goal position while avoiding two randomly-placed circular obstacles.

\paragraph{State space} (14-dimensional): See Table~\ref{tab:pm_obs}.

\begin{table}[htbp]
  \centering
  \caption{PointMassNavigate observation space.}
  \label{tab:pm_obs}
  \begin{tabular}{@{}lll@{}}
    \toprule
    \textbf{Component} & \textbf{Dimensions} & \textbf{Description} \\
    \midrule
    Position       & 2 & $(x, y) \in [-1, 1]^2$ \\
    Velocity       & 2 & $(v_x, v_y)$ \\
    Goal position  & 2 & Target $(x_g, y_g)$ \\
    Last action    & 2 & Previous force command \\
    Obstacle 1     & 3 & Position $(x_1, y_1)$ + radius $r_1$ \\
    Obstacle 2     & 3 & Position $(x_2, y_2)$ + radius $r_2$ \\
    \bottomrule
  \end{tabular}
\end{table}

\paragraph{Action space:} $\cA = [-1, 1]^2$ (2D force vector, scaled by 5.0\,N).

\paragraph{Dynamics} (semi-implicit Euler, $\Delta t = 1/60$\,s):
\begin{align}
  \bvf &\leftarrow \bvf \cdot (1 - c_d \Delta t) \\
  \bvf &\leftarrow \bvf + \frac{\bff}{m} \Delta t \\
  \bpf &\leftarrow \text{clamp}\!\left(\bpf + \bvf \Delta t,\; [-1, 1]^2\right)
\end{align}
where $m = 1.0$\,kg, $c_d = 5.0$, $\bff = \ba \times 5.0$\,N.

\paragraph{Reward:}
\begin{equation}
  r = -\|\bpf - \bpf_g\| + 0.5\exp\!\left(-\frac{\|\bpf - \bpf_g\|^2}{2 \cdot 0.05^2}\right) - 0.01\|\bvf\|^2 - 0.001\|\ba - \ba_{\text{prev}}\|^2 - 2.0 \sum_{i=1}^{2} \max(0, r_i - d_i)^2
  \label{eq:pm_reward}
\end{equation}

\paragraph{Success condition:} $\|\bpf - \bpf_g\| < 0.03$\,m.

\paragraph{Episode length:} 100 steps (default).

\paragraph{APG gradient path:} $\ba \rightarrow \bff \rightarrow \bvf \rightarrow \bpf \rightarrow r$ (fully differentiable Euler integration).

\subsection{PushT}
\label{sec:env_pusht}

\paragraph{Task:} Push a T-shaped rigid block from a random initial pose to a random goal pose using external forces and torques.

\paragraph{State space} (9-dimensional): See Table~\ref{tab:pt_obs}.

\begin{table}[htbp]
  \centering
  \caption{PushT observation space.}
  \label{tab:pt_obs}
  \begin{tabular}{@{}lll@{}}
    \toprule
    \textbf{Component} & \textbf{Dimensions} & \textbf{Description} \\
    \midrule
    Block position        & 2 & $(x, y) \in [-0.25, 0.25]^2$ \\
    Block angle           & 1 & $\theta \in [-\pi, \pi]$ \\
    Goal position         & 2 & Target $(x_g, y_g)$ \\
    Goal angle            & 1 & Target $\theta_g$ \\
    Block velocity        & 2 & $(v_x, v_y)$ \\
    Block angular velocity & 1 & $\omega$ \\
    \bottomrule
  \end{tabular}
\end{table}

\paragraph{Action space:} $\cA = [-1, 1]^3$ (force $x$, force $y$, torque $z$).

\paragraph{T-block geometry:} Top bar: $0.12 \times 0.03$\,m. Stem: $0.03 \times 0.09$\,m. Total height $\approx 0.12$\,m.

\paragraph{Dynamics} (2D rigid body Euler, $\Delta t = 1/60$\,s):
\begin{align}
  \bvf &\leftarrow \bvf \cdot (1 - c_d \Delta t), & \omega &\leftarrow \omega \cdot (1 - c_\omega \Delta t) \\
  \bvf &\leftarrow \bvf + \frac{\bff}{m} \Delta t,  & \omega &\leftarrow \omega + \frac{\tau}{I} \Delta t \\
  \bpf &\leftarrow \text{clamp}\!\left(\bpf + \bvf \Delta t,\; [-0.25, 0.25]^2\right), & \theta &\leftarrow \text{wrap}(\theta + \omega \Delta t)
\end{align}
where $m = 0.1$\,kg, $I = 1.95 \times 10^{-4}$\,kg$\cdot$m$^2$, $c_d = 5.0$, $c_\omega = 3.0$, $\bff = \ba_{[0:2]} \times 5.0$\,N, $\tau = a_{[2]} \times 0.2$\,Nm.

\paragraph{Reward:}
\begin{equation}
  r = -\|\bpf - \bpf_g\| + 0.5\exp\!\left(-\frac{\|\bpf - \bpf_g\|^2}{2 \cdot 0.02^2}\right) - 0.5\,|\text{wrap}(\theta - \theta_g)| - 0.001\,\|\ba - \ba_{\text{prev}}\|^2
  \label{eq:pt_reward}
\end{equation}

\paragraph{Success condition:} $\|\bpf - \bpf_g\| < 0.01$\,m and $|\text{wrap}(\theta - \theta_g)| < 0.1$\,rad.

\paragraph{Episode length:} 100 steps (default).

\paragraph{APG gradient path:} $\ba \rightarrow (\bff, \tau) \rightarrow (\bvf, \omega) \rightarrow (\bpf, \theta) \rightarrow r$.

\subsection{FrankaReach}
\label{sec:env_franka}

\paragraph{Task:} Control a 7-DOF Franka FR3 manipulator to reach a target 6D end-effector pose (position + orientation).

\paragraph{State space} (28-dimensional): See Table~\ref{tab:fr_obs}.

\begin{table}[htbp]
  \centering
  \caption{FrankaReach observation space.}
  \label{tab:fr_obs}
  \begin{tabular}{@{}lll@{}}
    \toprule
    \textbf{Component} & \textbf{Dimensions} & \textbf{Description} \\
    \midrule
    Joint positions     & 7 & $q_1, \ldots, q_7$ (arm joints) \\
    EEF position        & 3 & $(x, y, z)$ of end-effector \\
    EEF quaternion      & 4 & $(x, y, z, w)$ orientation \\
    Target position     & 3 & Goal $(x_g, y_g, z_g)$ \\
    Target quaternion   & 4 & Goal orientation $(x_g, y_g, z_g, w_g)$ \\
    Last action         & 7 & Previous joint velocity command \\
    \bottomrule
  \end{tabular}
\end{table}

\paragraph{Action space:} $\cA = [-1, 1]^7$ (delta joint positions, scaled by 0.2\,rad).

\paragraph{Dynamics:} Newton Physics engine with forward kinematics. Joint targets are computed as $q_{\text{target}} = \text{clamp}(q_{\text{current}} + \ba \times 0.2, \; q_{\text{min}}, q_{\text{max}})$.

\paragraph{Target sampling:} Positions sampled uniformly in $x \in [0.05, 0.70]$, $y \in [-0.45, 0.45]$, $z \in [0.20, 0.95]$\,m. Orientations sampled with tilt up to $\pm 60^\circ$ from the default downward pose.

\paragraph{Reward:}
\begin{equation}
  r = -0.2\,\|\bpf_{\text{eef}} - \bpf_{\text{target}}\| + 0.1\exp\!\left(-\frac{\|\bpf_{\text{eef}} - \bpf_{\text{target}}\|^2}{2 \cdot 0.1^2}\right) - 0.1 \cdot d_{\text{quat}}^2(\mathbf{q}_{\text{eef}}, \mathbf{q}_{\text{target}}) - 0.0001\,\|\ba - \ba_{\text{prev}}\|^2
  \label{eq:fr_reward}
\end{equation}

\paragraph{Success condition:} $\|\bpf_{\text{eef}} - \bpf_{\text{target}}\| < 0.005$\,m and $d_{\text{quat}}^2(\mathbf{q}_{\text{eef}}, \mathbf{q}_{\text{target}}) < 0.1$, where $d_{\text{quat}}^2(\mathbf{q}_1, \mathbf{q}_2) = \min(\|\mathbf{q}_1 - \mathbf{q}_2\|^2,\,\|\mathbf{q}_1 + \mathbf{q}_2\|^2)$ handles the quaternion double cover.

\paragraph{Episode length:} 30 steps (default).

\paragraph{APG gradient path:} $\ba \rightarrow q_{\text{target}} = \text{clamp}(q + \ba \times 0.2) \rightarrow \text{FK}(q_{\text{target}}) \rightarrow r$ (reward differentiable via Warp tape); $\ba \rightarrow q_{\text{target}}$ also constructed in PyTorch for differentiable joint-position observations.

The gradient computation uses a custom \texttt{torch.autograd.Function} that bridges the Warp-tape autodiff in Newton Physics with PyTorch's autograd system (see Section~\ref{sec:gradient_bridge}).

\subsection{Environment Complexity Comparison}
\label{sec:env_comparison}

Table~\ref{tab:env_comparison} compares the four environments across key properties.

\begin{table}[htbp]
  \centering
  \small
  \caption{Comparison of environment properties.}
  \label{tab:env_comparison}
  \begin{tabularx}{\textwidth}{@{}lXXXX@{}}
    \toprule
    \textbf{Property} & \textbf{PointMassSimple} & \textbf{PointMass} & \textbf{PushT} & \textbf{FrankaReach} \\
    \midrule
    Observation dim        & 3                    & 14                        & 9                          & 28 \\
    Action dim             & 1                    & 2                         & 3                          & 7 \\
    Dynamics               & 1D point mass (Euler)& 2D Point Mass (Euler)        & 2D rigid body (Euler)      & Articulated body (FK) \\
    DOF                    & 1 (pos)              & 2 (pos)                   & 3 (pos + rot)              & 7 (joints) \\
    Max episode steps      & 200                  & 100                       & 100                        & 30 \\
    Gradient bridge        & Native PyTorch       & Native PyTorch            & Native PyTorch             & Warp $\leftrightarrow$ PyTorch \\
    Obstacles/constraints  & None                 & 2 circles                 & Workspace bounds           & Joint limits \\
    Orientation            & No                   & No                        & Yes ($\theta$)             & Yes (quaternion) \\
    \bottomrule
  \end{tabularx}
\end{table}

\section{Experiments}
\label{sec:experiments}

\subsection{Experimental Setup}
\label{sec:setup}

\paragraph{Shared hyperparameters} (identical for PPO and APG unless noted): See Table~\ref{tab:shared_hp}.

\begin{table}[htbp]
  \centering
  \caption{Shared hyperparameters.}
  \label{tab:shared_hp}
  \begin{tabular}{@{}ll@{}}
    \toprule
    \textbf{Hyperparameter} & \textbf{Value} \\
    \midrule
    Learning rate                   & $2.5 \times 10^{-4}$ \\
    LR schedule                     & Linear annealing to 0 \\
    Discount factor $\gamma$        & 0.99 \\
    Max gradient norm               & 0.5 \\
    Optimizer                       & Adam ($\epsilon = 10^{-5}$) \\
    Observation normalization       & Welford running mean/std \\
    Number of parallel envs         & 4 \\
    Network architecture            & 2-layer MLP (64 units, Tanh, no LayerNorm) \\
    Total environment steps         & 500{,}000 \\
    \bottomrule
  \end{tabular}
\end{table}

\paragraph{PPO-specific hyperparameters:} See Table~\ref{tab:ppo_hp}.

\begin{table}[htbp]
  \centering
  \caption{PPO-specific hyperparameters.}
  \label{tab:ppo_hp}
  \begin{tabular}{@{}ll@{}}
    \toprule
    \textbf{Hyperparameter} & \textbf{Value} \\
    \midrule
    Steps per rollout           & 128 \\
    Minibatches                 & 4 \\
    Update epochs               & 4 \\
    GAE $\lambda$               & 0.95 \\
    Clip coefficient $\epsilon$ & 0.2 \\
    Value function coefficient  & 0.5 \\
    Entropy coefficient         & 0.01 \\
    Clipped value loss          & Yes \\
    Advantage normalization     & Yes \\
    \bottomrule
  \end{tabular}
\end{table}

\paragraph{APG-specific hyperparameters:} See Table~\ref{tab:apg_hp}.

\begin{table}[htbp]
  \centering
  \caption{APG-specific hyperparameters.}
  \label{tab:apg_hp}
  \begin{tabular}{@{}ll@{}}
    \toprule
    \textbf{Hyperparameter} & \textbf{Value} \\
    \midrule
    Gradient steps per iteration   & 8 \\
    Segment length                 & Full episode \\
    Bootstrap mode                 & Critic \\
    \bottomrule
  \end{tabular}
\end{table}

\subsection{Evaluation Protocol}
\label{sec:eval_protocol}

We employ a rigorous evaluation protocol to ensure fair comparison:

\begin{enumerate}
  \item \textbf{Deterministic evaluation:} At regular intervals, we run a fixed number of evaluation episodes in a separate black-box environment using the greedy policy: argmax over logits for discrete action spaces, or the clamped actor mean $\text{clip}(\bmu_{\btheta}(s), -1, 1)$ for continuous spaces. Observations are normalized via the shared running normalizer (no update during eval). The following metrics are collected per evaluation checkpoint:
  \begin{itemize}
    \item \textbf{Episodic return}: mean undiscounted return over completed episodes.
    \item \textbf{Episodic length}: mean number of steps per episode.
    \item \textbf{Success rate}: fraction of episodes that terminate by reaching the goal, as opposed to timeout truncation.
    \item \textbf{Mean time to success}: mean episode length conditioned on successful episodes only.
  \end{itemize}

  \item \textbf{Multi-axis logging:} Metrics are recorded against two independent axes to enable fair comparison between PPO and APG:
  \begin{itemize}
    \item \textbf{Environment steps}: Total number of environment interactions (measures sample efficiency).
    \item \textbf{Gradient steps}: Total number of optimizer updates (measures compute efficiency, equalized across algorithms for controlled comparison).
  \end{itemize}

  \item \textbf{Seed sweep:} Results are reported as mean $\pm$ std over multiple random seeds.
\end{enumerate}

\paragraph{Counting environment steps.}
To ensure the environment-step axis is comparable across algorithms, we count one environment step as one call to the environment \texttt{step()}---that is, one transition $(s_t, a_t, r_t, s_{t+1})$---irrespective of whether that transition participates in a differentiable computation graph. For PPO, steps accumulate over on-policy rollouts across the parallel environments. For APG, the stateful rollout paradigm (Section~\ref{sec:training_algorithms}) carries environment state across gradient steps, but every executed action still produces exactly one counted transition; resets triggered by episode termination or truncation are not charged to the budget. Because PPO performs multiple update epochs over a stored batch while APG rolls out fresh trajectories per gradient step, the two algorithms consume environment steps at different rates for a fixed number of gradient updates---precisely the confound that motivates reporting the two axes independently.

\subsection{Experiment Results}
\label{sec:results}

\begin{figure}[htbp]
  \centering
  \begin{subfigure}[b]{0.48\textwidth}
    \includegraphics[width=\textwidth]{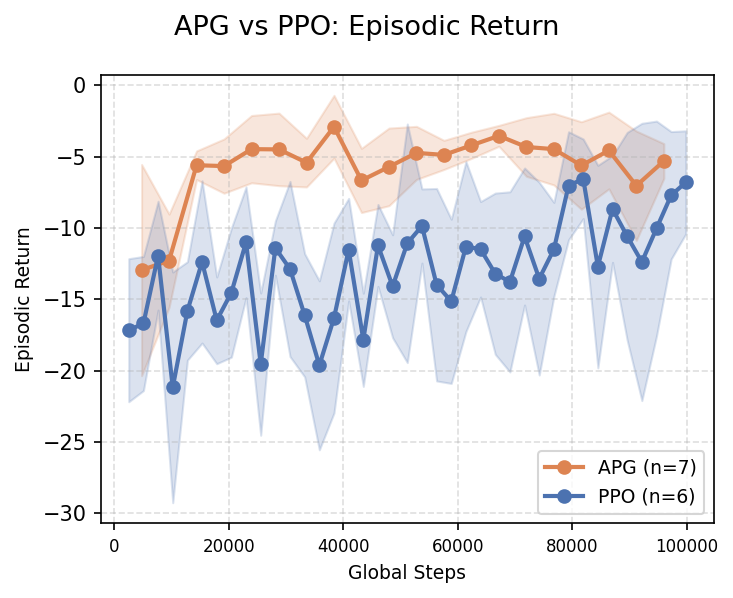}
    \caption{PointMassSimple}
  \end{subfigure}
  \hfill
  \begin{subfigure}[b]{0.48\textwidth}
    \includegraphics[width=\textwidth]{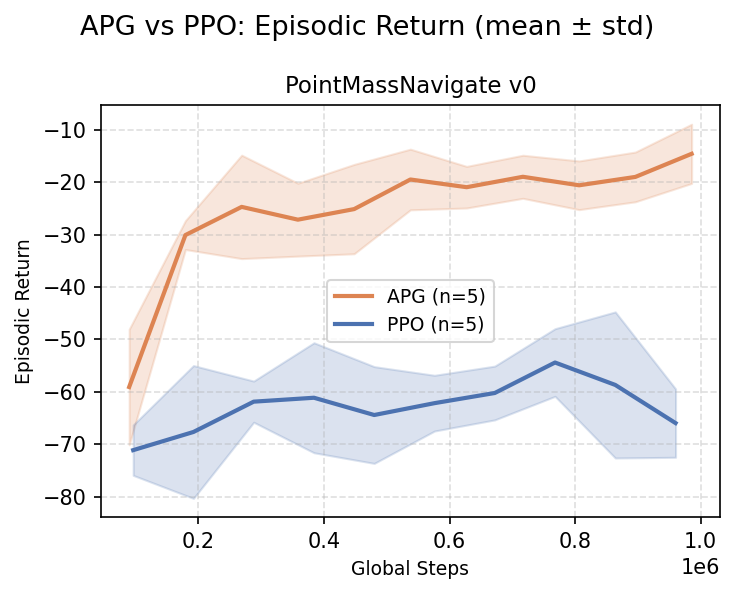}
    \caption{PointMassNavigate}
  \end{subfigure}

  \vspace{0.5em}

  \begin{subfigure}[b]{0.48\textwidth}
    \includegraphics[width=\textwidth]{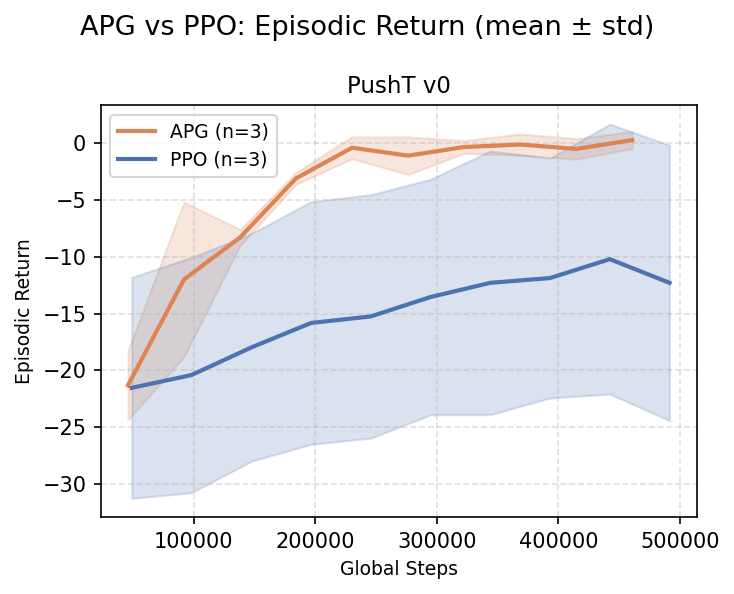}
    \caption{PushT}
  \end{subfigure}
  \hfill
  \begin{subfigure}[b]{0.48\textwidth}
    \includegraphics[width=\textwidth]{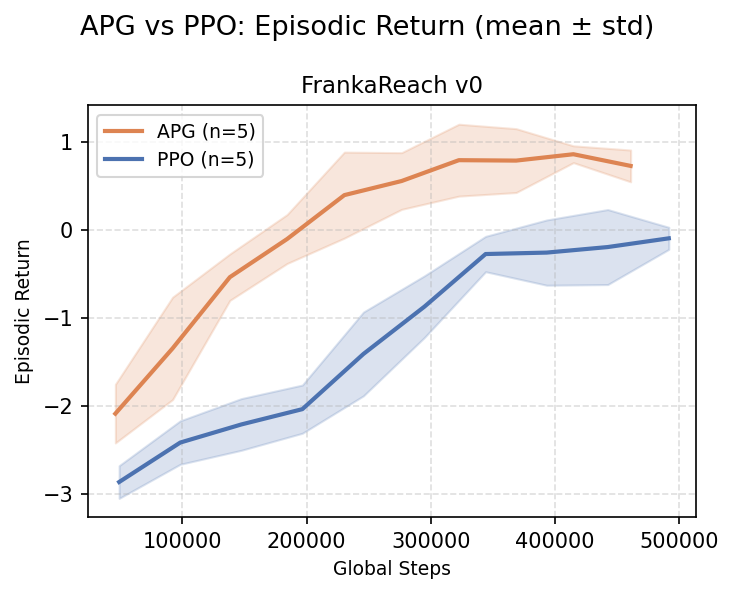}
    \caption{FrankaReach}
  \end{subfigure}
  \caption{Episodic return (mean $\pm$ std) vs.\ total gradient steps for PPO (blue) and APG (orange). The x-axis is equalized: both algorithms execute the same total number of gradient updates.}
  \label{fig:return_curves}
\end{figure}

Figure~\ref{fig:return_curves} shows the episodic return curves for all four environments, plotted against total gradient steps to enable a fair compute-controlled comparison between PPO and APG; the companion environment-step axis is summarized in Table~\ref{tab:sample_eff}. Table~\ref{tab:final_return} summarises the final returns.

\paragraph{Performance thresholds and success ratio.}
The efficiency tables (Tables~\ref{tab:sample_eff}--\ref{tab:grad_eff}) report the number of environment/gradient steps at which the evaluation episodic return first crosses an environment-specific performance threshold: $-10$ for PointMassNavigate, $-1$ for PushT, and $0.5$ for FrankaReach (PointMassSimple is near-solved by both methods and is reported at full success). The ``Success Ratio'' column reports the final deterministic-evaluation goal-reaching success rate of Section~\ref{sec:eval_protocol}---the fraction of evaluation episodes that terminate at the goal rather than by timeout---which is a strictly stronger criterion than the return threshold: a policy may attain high dense reward yet still fail to satisfy the tight pose/position tolerance required for success. This distinction is important for interpreting the results: for example, on FrankaReach the APG policy achieves a substantially higher dense return than PPO (Table~\ref{tab:final_return}) even though the strict goal-reaching success rate remains partial, because the reward shaping rewards proximity rather than exact pose matching. When an algorithm does not cross the return threshold within its training budget, we report the total budget consumed.

\begin{table}[htbp]
  \centering
  \caption{Final performance (episodic return). Results reported as mean $\pm$ std over multiple seeds.}
  \label{tab:final_return}
  \begin{tabular}{@{}lcc@{}}
    \toprule
    \textbf{Environment} & \textbf{PPO (mean $\pm$ std)} & \textbf{APG (mean $\pm$ std)} \\
    \midrule
    PointMassSimple & $-7.14 \pm 3.87$ & $-5.66 \pm 1.28$ \\
    PointMassNavigate & $-64.89 \pm 11.92$ & $-14.59 \pm 3.36$ \\
    PushT             & $-12.29 \pm 12.13$   & $0.29 \pm 0.76$ \\
    FrankaReach       & $-0.09 \pm 0.13$    & $0.73 \pm 0.08$ \\
    \bottomrule
  \end{tabular}
\end{table}

\begin{table}[htbp]
  \centering
  \caption{Sample efficiency: environment steps to reach performance threshold.}
  \label{tab:sample_eff}
  \begin{tabular}{@{}lcrrr@{}}
    \toprule
    \textbf{Environment} & \textbf{Success Ratio} & \textbf{PPO Steps} & \textbf{APG Steps} & \textbf{APG Reduction} \\
    \midrule
    PointMassSimple & 100\% & 69120 & 38400 & 1.80x \\
    PointMassNavigate & 20\% & 263360 & 192000 & 1.37x \\
    PushT             & 100\% & 147456 & 138240 & 1.07x \\
    FrankaReach       & 20\% & 1000000 & 230400 & 4.34x \\
    \bottomrule
  \end{tabular}
\end{table}

\begin{table}[htbp]
  \centering
  \caption{Learning efficiency: gradient steps to reach performance threshold.}
  \label{tab:grad_eff}
  \begin{tabular}{@{}lcrrr@{}}
    \toprule
    \textbf{Environment} & \textbf{Success Ratio} & \textbf{PPO Grad Steps} & \textbf{APG Grad Steps} & \textbf{APG Speedup} \\
    \midrule
    PointMassSimple & 100\% & 2150 & 320 & 6.72x \\
    PointMassNavigate & 20\% & 1104 & 120 & 9.2x \\
    PushT             & 100\% & 560 & 144 & 3.89x \\
    FrankaReach       & 20\% & 3824 & 240 & 15.93x \\
    \bottomrule
  \end{tabular}
\end{table}

\subsection{Ablation: Bootstrap Mode for Segmented APG}
\label{sec:ablation_bootstrap}

We conduct an ablation study on PointMassNavigate to isolate the effect of the bootstrap strategy in segmented APG (segment length $L=25$, horizon $T=100$, giving 4 segments per episode). Two modes are compared across 3 seeds each:

\begin{itemize}
  \item \textbf{MC bootstrap}: future returns for each segment are pre-computed by backward accumulation of detached rewards from all subsequent segments, properly discounted by $\gamma^{L_k}$ at each segment boundary.
  \item \textbf{Critic bootstrap}: a learned value function $V_\phi(s)$ estimates the discounted future return at each segment boundary ($b_k = \gamma^{L_k} V_\phi(s_{k,\text{end}})$). The critic is trained concurrently via MSE regression to segment-level bootstrapped returns. Crucially, $s_{k,\text{end}}$ is kept in the computation graph (not detached), so $\nabla_{\btheta} b_k$ extends the effective gradient horizon through the critic to the terminal state of each segment.
\end{itemize}

Both modes reach a success rate of $1.0$ by the end of training. The critic bootstrap converges to a slightly higher final episodic return ($-14.44 \pm 3.38$) than MC bootstrap ($-15.31 \pm 2.84$), though the difference is within one standard deviation given three seeds. The critic bootstrap exhibits marginally faster early-phase improvement, consistent with the expected benefit of a learned value function reducing variance in the return estimate at segment boundaries.

\begin{figure}[htbp]
  \centering
  \begin{subfigure}[b]{0.48\textwidth}
    \includegraphics[width=\textwidth]{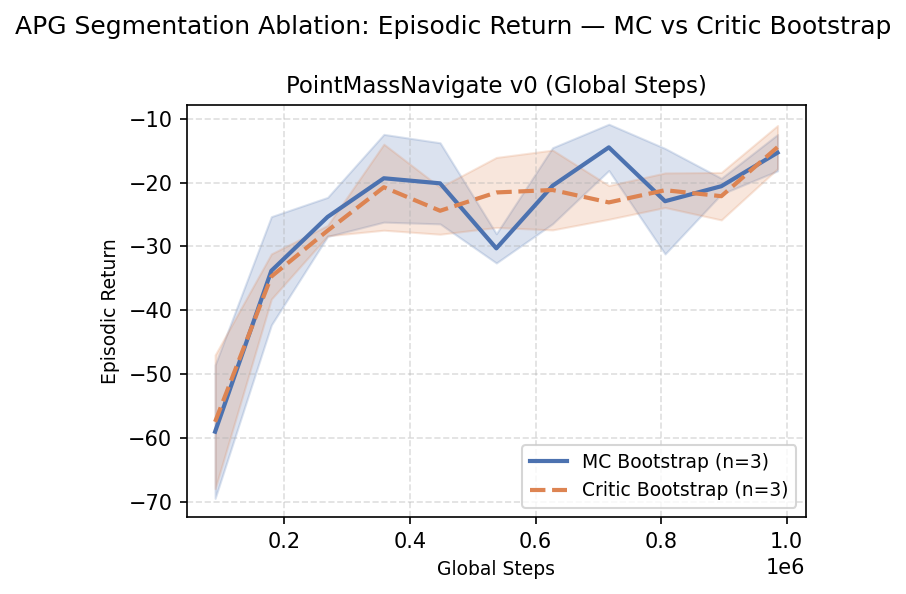}
    \caption{Return vs.\ environment steps}
  \end{subfigure}
  \hfill
  \begin{subfigure}[b]{0.48\textwidth}
    \includegraphics[width=\textwidth]{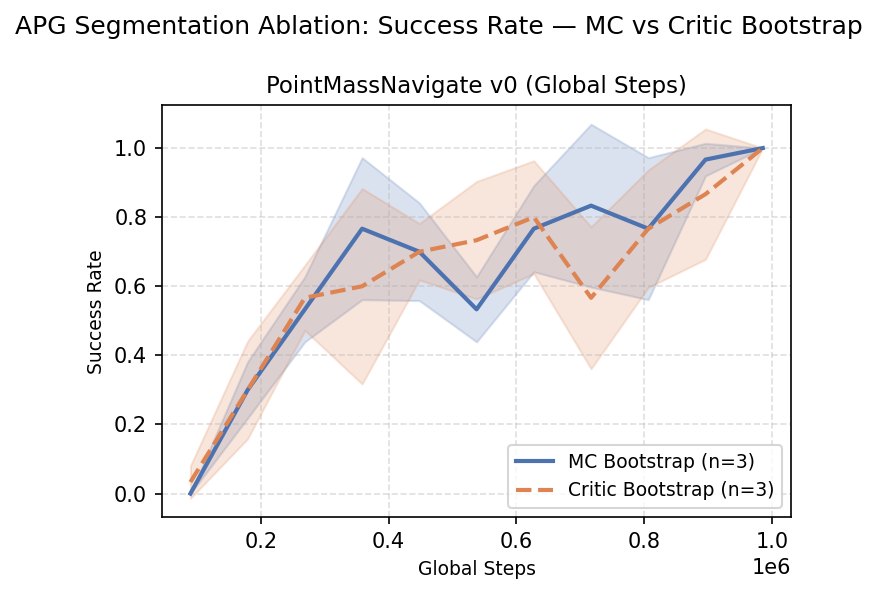}
    \caption{Success rate vs.\ environment steps}
  \end{subfigure}
  \caption{Bootstrap-mode ablation on PointMassNavigate.}
  \label{fig:ablation_bootstrap}
\end{figure}

\subsection{Ablation: Segment Length with MC and Critic Bootstrap}
\label{sec:ablation_segment}

To understand how segment length $L$ interacts with the bootstrap strategy, we sweep $L \in \{10, 25, 50\}$ under both MC and critic bootstrap on PointMassNavigate (horizon $T=100$, one seed per condition). This isolates the credit-assignment range—shorter segments truncate gradient chains more aggressively, relying more heavily on the bootstrap estimate to carry future-return information across boundaries.

Table~\ref{tab:ablation_segment} summarizes the results.

\textbf{MC bootstrap} is robust across all three segment lengths. Even at $L=10$ (10 segments per episode) it achieves a success rate of 0.9 and converges to a reasonable final return of $-22.3$. Performance improves monotonically with $L$: at $L=25$ and $L=50$ the agent reaches a success rate of 1.0, with final returns of $-15.1$ and $-10.5$ respectively. The consistent gradient signal from pre-computed discounted returns makes MC bootstrap tolerant of shorter horizons.

\textbf{Critic bootstrap} exhibits a strong sensitivity to $L$. At $L=10$, the critic fails to provide reliable bootstrap targets early in training, and the agent converges to a degenerate policy with zero success rate and a return of $-61.7$. At $L=25$ the critic partially recovers (success rate 0.5, return $-34.5$), and at $L=50$ it matches MC bootstrap ($-9.8$, success 1.0). The critic's value estimates require a sufficient gradient horizon to stabilize: when segments are too short, the critic's bootstrapped targets are too noisy to extend effective credit assignment beyond the segment boundary, and the additional critic loss introduces conflicting gradients.

The results suggest that MC bootstrap should be preferred as the default for short-to-medium segment lengths, while critic bootstrap becomes competitive only when $L$ is large enough to supply stable training signal to the value network.

\begin{figure}[htbp]
  \centering
  \begin{subfigure}[b]{0.48\textwidth}
    \includegraphics[width=\textwidth]{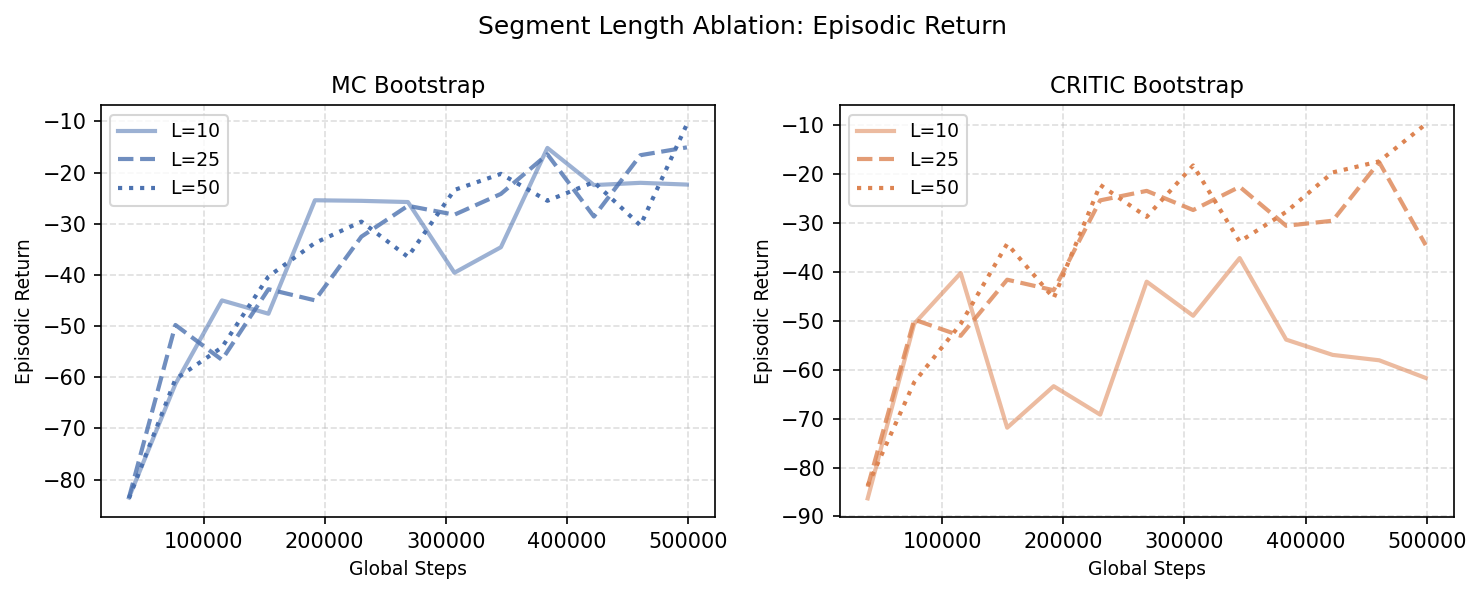}
    \caption{Final return}
  \end{subfigure}
  \hfill
  \begin{subfigure}[b]{0.48\textwidth}
    \includegraphics[width=\textwidth]{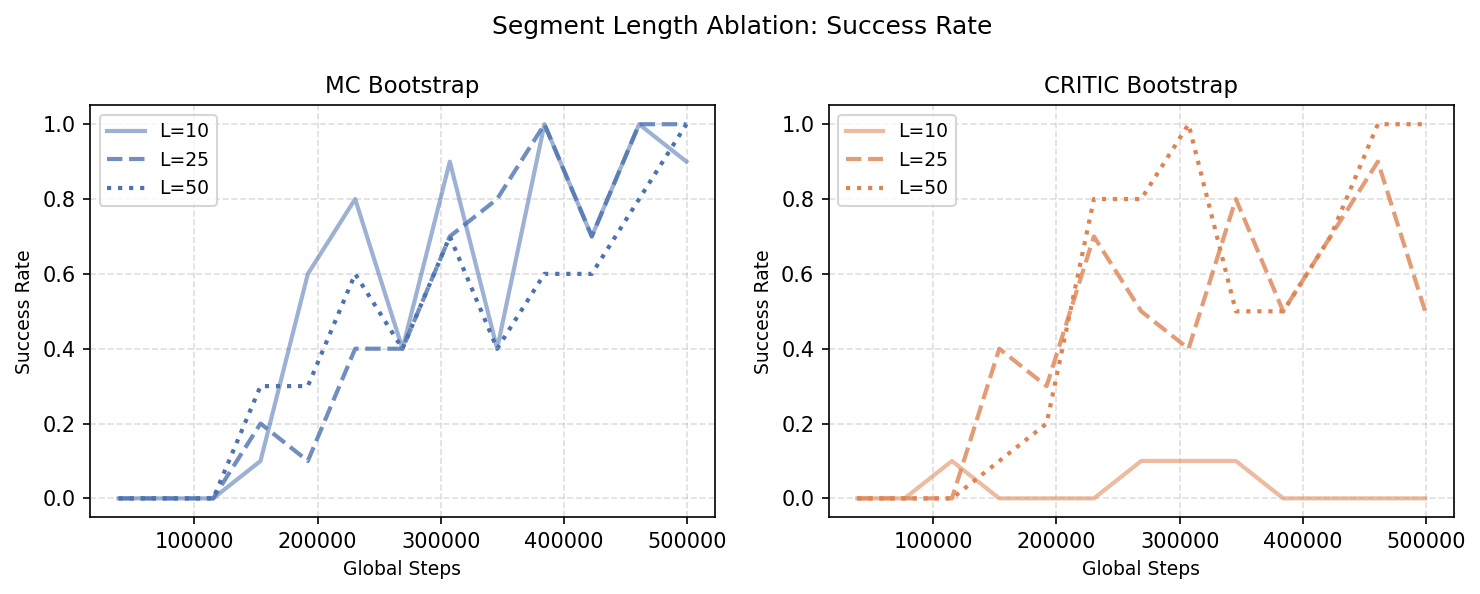}
    \caption{Success rate}
  \end{subfigure}
  \caption{Segment-length ablation on PointMassNavigate.}
  \label{fig:ablation_segment}
\end{figure}

\begin{table}[htbp]
  \centering
  \caption{Segment length ablation: final episodic return and success rate on PointMassNavigate (1 seed per condition, 500K steps).}
  \label{tab:ablation_segment}
  \begin{tabular}{@{}llrr@{}}
    \toprule
    \textbf{Bootstrap} & \textbf{$L$} & \textbf{Final Return} & \textbf{Success Rate} \\
    \midrule
    MC     & 10 & $-22.3$ & 0.90 \\
    MC     & 25 & $-15.1$ & 1.00 \\
    MC     & 50 & $-10.5$ & 1.00 \\
    \midrule
    Critic & 10 & $-61.7$ & 0.00 \\
    Critic & 25 & $-34.5$ & 0.50 \\
    Critic & 50 & $-9.8$  & 1.00 \\
    \bottomrule
  \end{tabular}
\end{table}

\subsection{Reproducibility}
\label{sec:reproducibility}
The full implementation, environment definitions, training scripts, and plotting utilities are open-sourced.\footnote{\url{https://github.com/yuecideng/analytic_policy_gradients}} All experiments use a single entry point, \texttt{rl.py}, selecting the algorithm via \texttt{-{}-algorithm \{ppo,apg\}} and the environment via \texttt{-{}-env\_id}; multi-seed sweeps are launched with \texttt{-{}-num\_seeds}, and the \texttt{-{}-equalize\_grad\_steps} flag aligns the total gradient-step budget between PPO and APG for the compute-controlled comparison. The exact commands reproducing each figure are provided in the repository \texttt{scripts/} directory: \texttt{plot\_return\_curves.py} (Fig.~\ref{fig:return_curves}), \texttt{plot\_bootstrap\_ablation.py} (Fig.~\ref{fig:ablation_bootstrap}), \texttt{plot\_segment\_ablation.py} (Fig.~\ref{fig:ablation_segment}), and \texttt{fetch\_results.py} (all result tables, parsed from local TensorBoard logs). Dependencies are listed in \texttt{requirements.txt}; the FrankaReach environment additionally requires \texttt{newton==1.1.0} and a CUDA-capable GPU. Runs are logged to TensorBoard and, optionally, Weights \& Biases (project \texttt{cleanRL}). The reported numbers correspond to commit \texttt{<fill-in>} evaluated on a single \texttt{<GPU model>} GPU.

\section{Discussion and Conclusion}
\label{sec:conclusion}

\subsection{Summary}
\label{sec:summary}

We presented a unified framework for comparing Analytic Policy Gradients (APG) and Proximal Policy Optimization (PPO) across four differentiable continuous control environments. Experiments on PointMassSimple, PointMassNavigate, PushT, and FrankaReach yield four concrete findings:

\begin{enumerate}
  \item \textbf{Gradient quality vs.\ variance:} APG computes exact policy gradients by backpropagating through dynamics, bypassing the variance inherent in Monte Carlo advantage estimation. Across the tested environments APG attains a higher final episodic return than PPO (Table~\ref{tab:final_return}) and reaches full goal-reaching success on the simpler tasks (PointMassSimple, PushT). On the harder tasks (PointMassNavigate, FrankaReach) the strict success rate remains partial even though the dense return improves substantially over PPO---a consequence of the gap between reward shaping and the tight pose tolerance required for success (Section~\ref{sec:results}). This confirms that exact gradients provide a reliable training signal while also delineating where dense-reward convergence does not by itself imply task success.

  \item \textbf{Sample efficiency:} APG requires substantially fewer gradient steps than PPO to reach comparable performance thresholds (Table~\ref{tab:grad_eff}): 6.7$\times$ fewer on PointMassSimple, 9.2$\times$ fewer on PointMassNavigate, 3.9$\times$ on PushT, and 15.9$\times$ on FrankaReach, demonstrating that the per-step quality of analytic gradients more than compensates for PPO's multi-epoch update schedule.

  \item \textbf{Segmentation and bootstrap strategy matter:} The segment length ablation (Section~\ref{sec:ablation_segment}) shows that MC bootstrap degrades gracefully as $L$ shrinks (success rate 0.9 at $L=10$), while critic bootstrap collapses entirely at $L=10$ (success rate 0.0) and recovers only at $L=50$. MC bootstrap is therefore the more robust default for short-to-medium horizons; critic bootstrap offers competitive final performance only when the segment length is large enough to stabilize its value targets.

  \item \textbf{Gradient bridge feasibility:} Our custom \texttt{torch.autograd.Function} demonstrates that analytic policy gradients are practical for GPU-accelerated physics engines that do not natively expose PyTorch-compatible derivatives, broadening APG beyond analytically specified environments.
\end{enumerate}

\subsection{Limitations}
\label{sec:limitations}

\begin{enumerate}
  \item \textbf{Environment differentiability requirement:} APG fundamentally requires differentiable environment dynamics. This limits applicability to simulations and excludes real-world training or environments with non-differentiable components (contacts, collisions with discrete events).

  \item \textbf{Gradient chain length:} Long episodes can lead to vanishing or exploding gradients despite segmentation. The effectiveness of APG may be sensitive to the choice of segment length and bootstrap strategy.

  \item \textbf{Compute overhead:} Maintaining the computation graph through environment rollouts incurs memory and compute overhead compared to PPO's detached rollouts, particularly for complex environments like FrankaReach.
\end{enumerate}

\subsection{Future Work}
\label{sec:future}

\begin{enumerate}

  \item \textbf{Hybrid APG--PPO algorithms with trust-region constraints.}
  The complementary strengths of APG and PPO suggest a natural synthesis: APG supplies low-variance analytic gradients, while PPO's clipped surrogate objective provides a conservative trust-region constraint that prevents catastrophic policy updates. Prior work on short-horizon actor-critic (SHAC) methods~\cite{guo2023shac} and differentiable simulation-based actors~\cite{xu2022accelerated} demonstrates that combining analytic rollouts with a learned value baseline can substantially stabilize training. A principled hybrid could apply the PPO ratio-clipping objective directly to the analytic gradient update, bounding effective policy step sizes while preserving gradient quality. Our unified implementation, which shares actor-critic architecture and observation normalization across both algorithms, is well-positioned to conduct a controlled investigation of this design space.

  \item \textbf{APG through learned differentiable world models.}
  The differentiability requirement currently restricts APG to analytically specified simulators. A natural generalization is to treat learned neural world models as the differentiable substrate, backpropagating policy gradients through model-predicted rollouts rather than ground-truth dynamics. Recent model-based RL systems such as DreamerV3~\cite{hafner2023dreamerv3} and TD-MPC2~\cite{hansen2024tdmpc2} demonstrate that latent imagination with differentiable recurrent transitions enables competitive policy optimization without environment rollouts. However, gradient compounding through approximate models introduces model-bias error that can dominate the variance reduction benefit. A systematic study comparing APG gradient quality under ground-truth versus learned dynamics---as a function of model capacity, rollout horizon, and Lipschitz constant---would clarify when learned-model gradients remain reliable signals and when model errors corrupt the optimization landscape.

  \item \textbf{Principled adaptive segment-length scheduling.}
  Our segment length ablation (Section~\ref{sec:ablation_segment}, Table~\ref{tab:ablation_segment}) reveals a clear bias--variance trade-off: shorter segments reduce memory overhead and gradient chain length but truncate credit assignment, and the impact is particularly severe for critic bootstrap. A theoretically principled curriculum could adapt $L_{\mathrm{seg}}$ online based on the gradient signal-to-noise ratio~\cite{pascanu2013exploding} or the convergence of the critic's bootstrap targets. The optimal segment length balances the bias introduced by truncated BPTT against gradient variance, an objective analogous to the $\lambda$ selection problem in GAE. Adaptive schedules grounded in this bias--variance decomposition, potentially implemented via meta-gradient methods that differentiate through the segment-length choice, would provide a more robust alternative to the fixed-length ablation studied here.

\end{enumerate}

\section*{References}
\label{sec:references}

\appendix

\section{Environment Visualizations}
\label{sec:appendix_vis}

Rendered frames and GIFs for the custom differentiable environments are produced by the environment-specific renderers during evaluation. PointMassSimple renders the one-dimensional track, target marker, success band, current position, and velocity arrow; PointMassNavigate, PushT, and FrankaReach use analogous task-specific visualizations.

\begin{figure}[htbp]
  \centering
  \begin{subfigure}[b]{0.45\textwidth}
    \includegraphics[width=\textwidth]{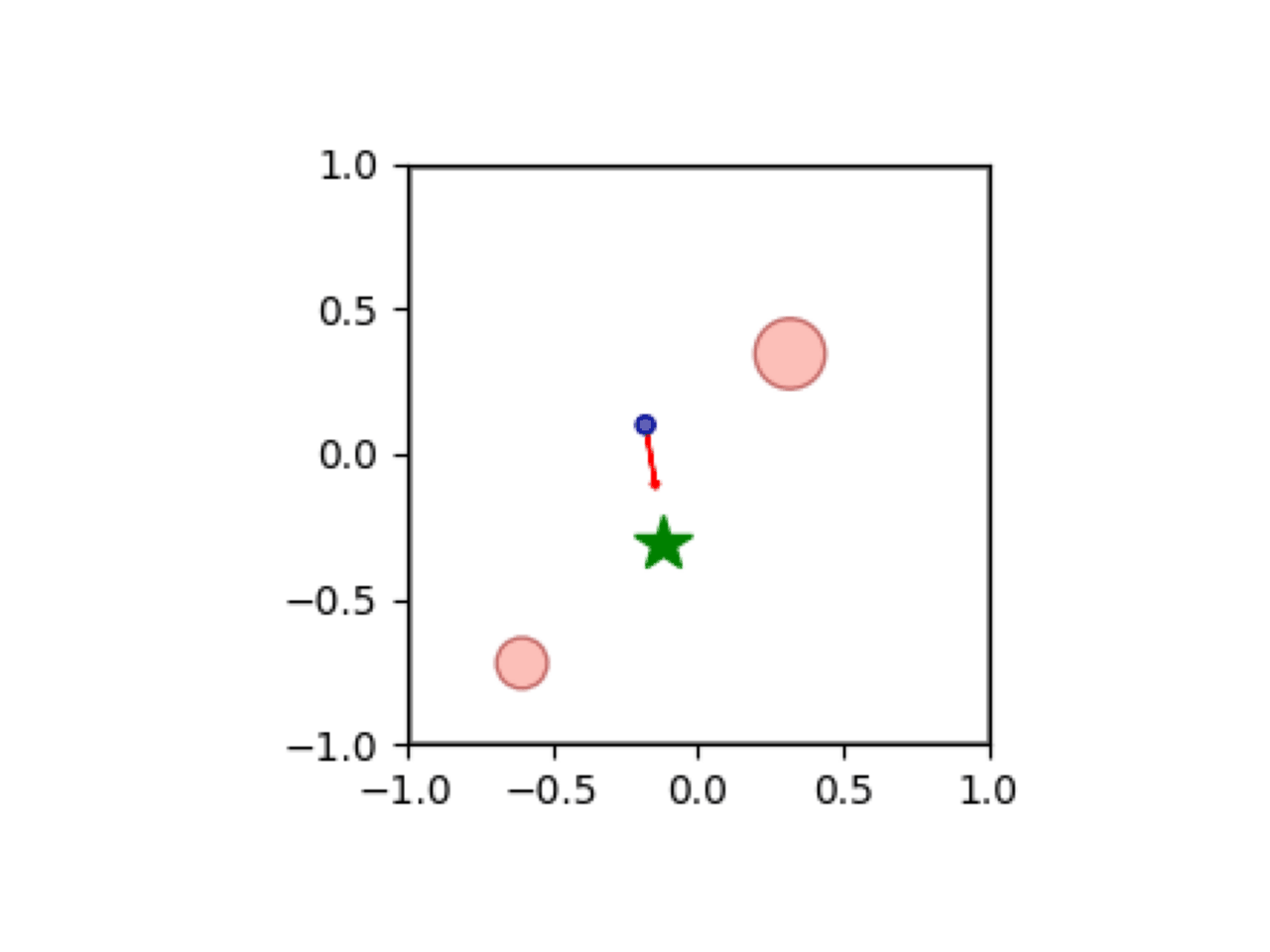}
    \caption{PointMassNavigate}
  \end{subfigure}
  \hfill
  \begin{subfigure}[b]{0.45\textwidth}
    \includegraphics[width=\textwidth]{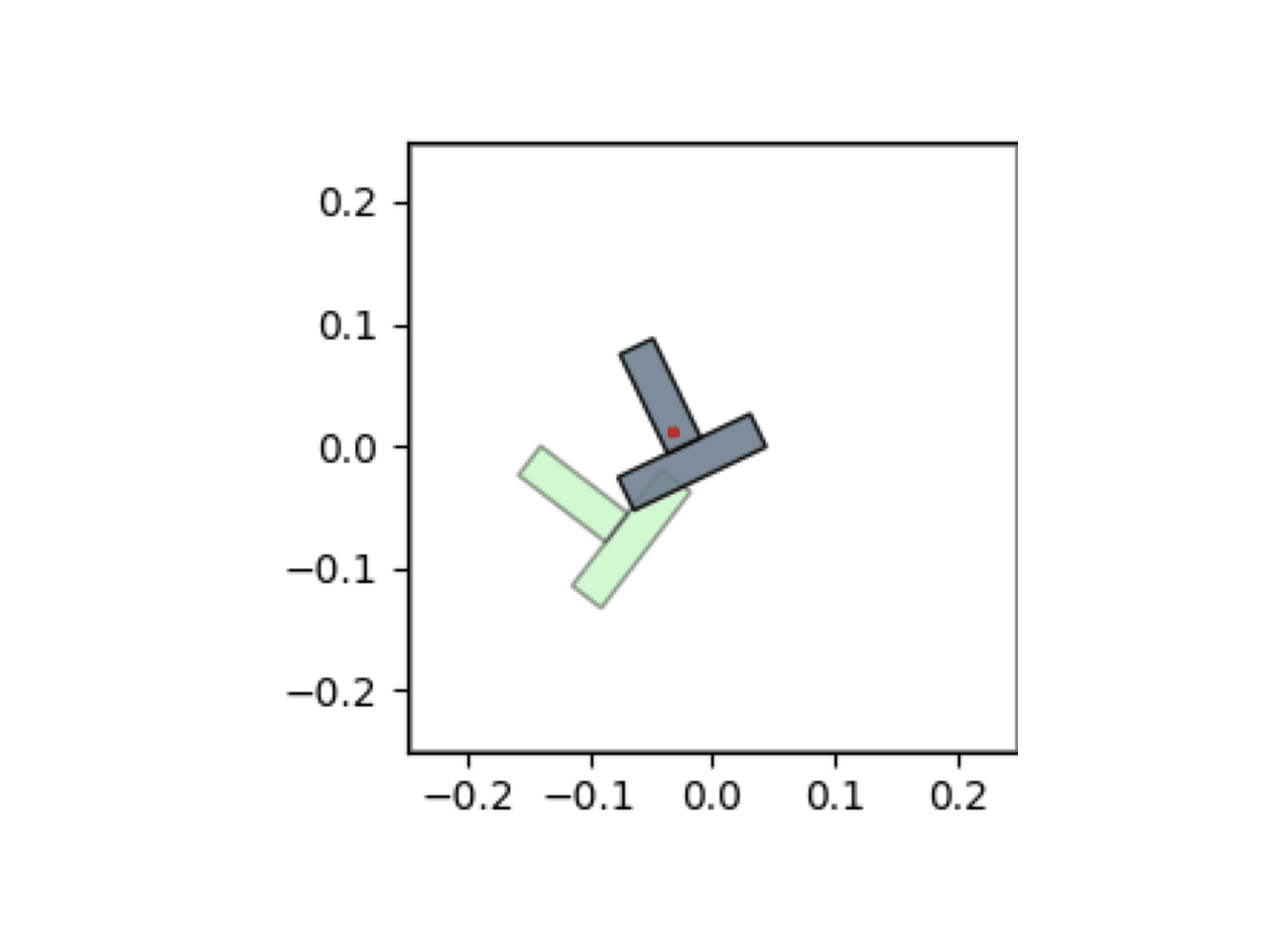}
    \caption{PushT}
  \end{subfigure}
  \vspace{0.3em}
  \begin{subfigure}[b]{0.45\textwidth}
    \includegraphics[width=\textwidth]{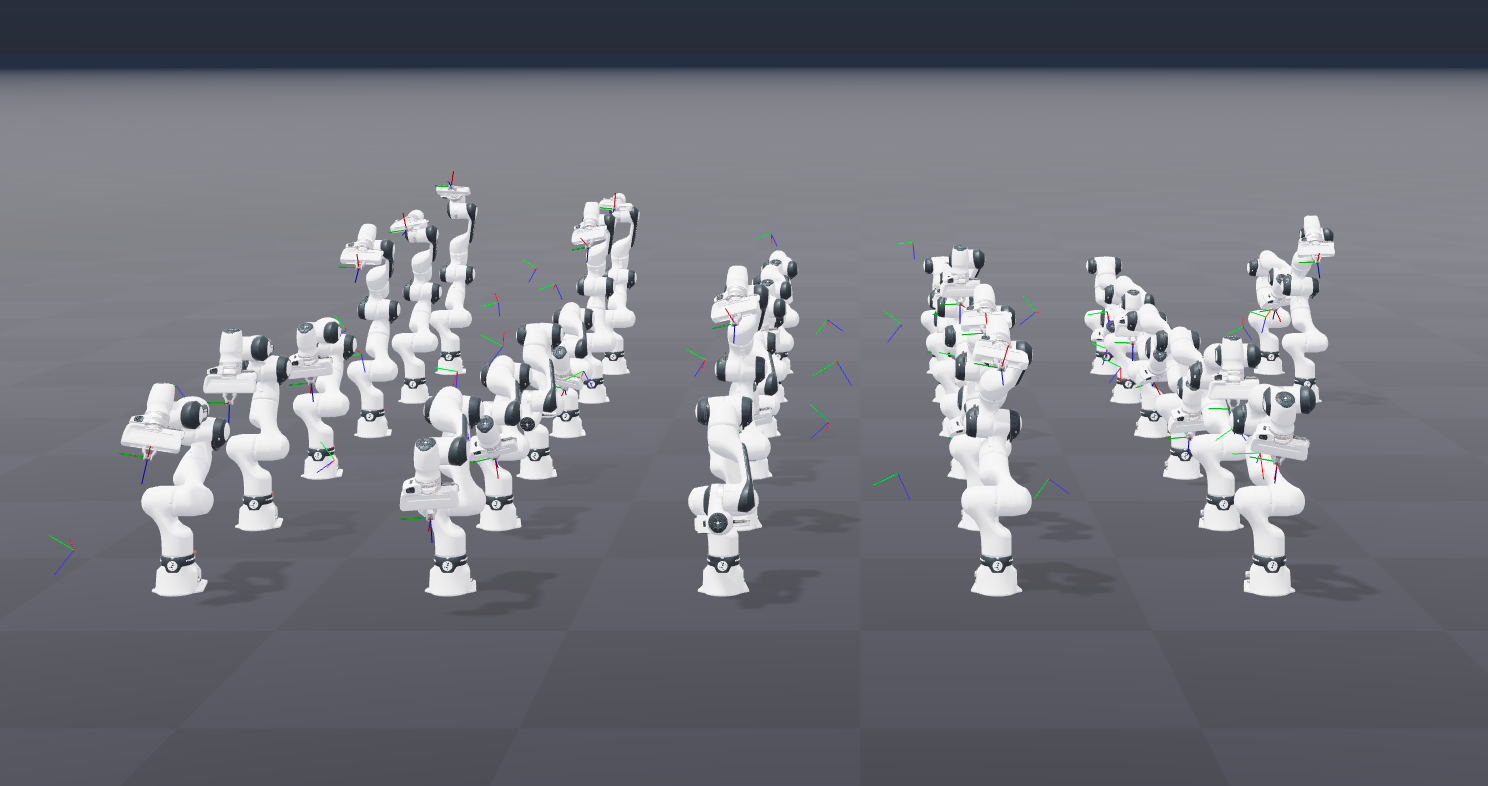}
    \caption{FrankaReach}
  \end{subfigure}
  \hfill
  \begin{subfigure}[b]{0.45\textwidth}
    \includegraphics[width=\textwidth]{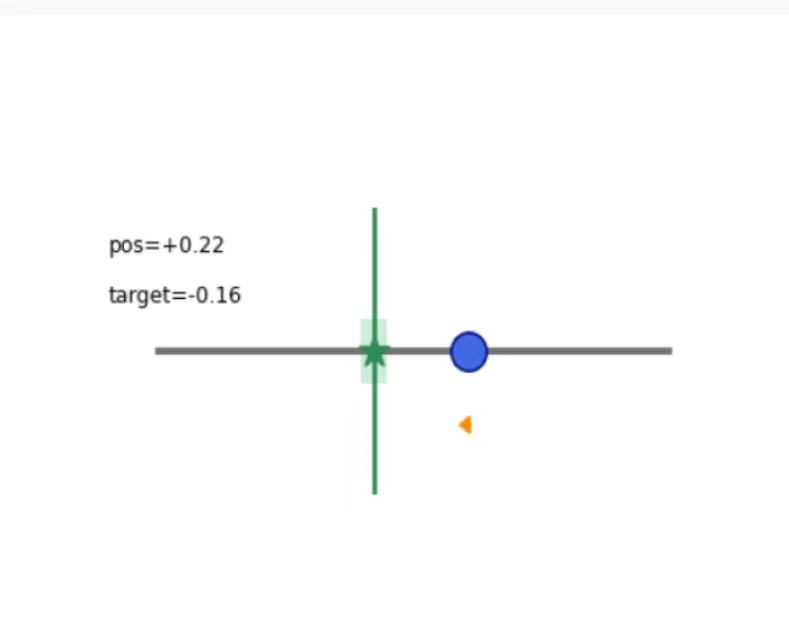}
    \caption{SimplePointMass}
  \end{subfigure}
  \caption{Environment renderings of the four continuous control tasks.}
  \label{fig:env_visualizations}
\end{figure}

\end{document}